\documentclass[11pt]{article}
\usepackage{acl2014}
\usepackage{amsmath}
\usepackage{times}
\usepackage{url}
\usepackage{latexsym}
\usepackage{color}
\usepackage{multicol}
\usepackage{array}
\usepackage{breqn}
\usepackage{algorithm}
\usepackage[noend]{algpseudocode}
\newfloat{algorithm}{t}{top}
\newcommand{\argmax}{\operatornamewithlimits{argmax}}

\newcommand{\dt}[1]{\left\langle #1 \right\rangle}
\newenvironment{LP}{\begin{equation}\vspace{-0.5em}\begin{array}{>{\displaystyle}r*4{>{\displaystyle}l}} }{\end{array}\end{equation}}

\newcommand{\segment}[2]{\textbf{[}#1\textbf{]}$_{\mbox{\tiny{#2}}}$}
\newcommand{\segmentm}[2]{#1$_{\mbox{\tiny{\textit{#2}}}}$}
\newcommand{\segments}[2]{#1$_{\mbox{\tiny{\textbf{#2}}}}$}

\newcommand{\blambda}{\lambda}

\title{Learning Soft Linear Constraints with Application to Citation Field Extraction}

\author{Sam Anzaroot \; Alexandre Passos \; David Belanger \; Andrew McCallum \\
  Department of Computer Science \\
  University of Massachusetts, Amherst \\
  {\tt \{anzaroot, apassos, belanger, mccallum\}@cs.umass.edu}
}

\date{}

\begin{document}
\maketitle
\begin{abstract}
Accurately segmenting a citation string into fields for authors, titles, etc. is a challenging  task because the output typically obeys various global constraints.  Previous work has shown that modeling  \textit{soft constraints}, where the model is encouraged, but not require to obey the constraints, can substantially improve segmentation performance. On the other hand, for imposing \textit{hard constraints}, dual decomposition is a popular technique for efficient prediction given existing algorithms for unconstrained inference. We extend dual decomposition to perform prediction subject to soft constraints. Moreover, with a technique for performing inference given soft constraints, it is easy to automatically generate large families of constraints and learn their costs with a simple convex optimization problem during training. This allows us to obtain substantial gains in accuracy on a new, challenging citation extraction dataset.
\end{abstract}

\section{Introduction}

	\begin{figure*}
          \begin{small}
            \segments{4 .}{ref-marker} \segment{ \segmentm{ \segments{
                  J. }{first} \segments{ D. }{middle} \segments{ Monk
                  ,}{last} }{person} }{authors} \segment{ Cardinal
              Functions on Boolean Algebra , }{title} \segment{
              \segmentm{ Lectures in Mathematics , ETH Zurich ,
              }{series} \segmentm{ Birkh¨ause Verlag , }{publisher}
              \segmentm{ Basel , Boston , Berlin , }{address}
              \segmentm{ \segments{ 1990 . }{year} }{date} }{venue}

          \end{small} 
          \caption{Example labeled citation}
	   \label{example}
 	\end{figure*}

Citation field extraction, an instance of information extraction, is the task of segmenting and labeling research paper citation strings into their constituent parts, including authors, editors, year, journal, volume, conference venue, etc.  This task is important because citation data is often provided only in plain text; however, having an accurate structured database of bibliographic information is necessary for many scientometric tasks, such as mapping scientific sub-communities, discovering research trends, and analyzing networks of researchers.  Automated citation field extraction needs further research because it has not yet reached a level of accuracy at which it can be practically deployed in real-world systems.

Hidden Markov models and linear-chain conditional random fields (CRFs) have previously been applied to citation extraction~\cite{hetzner2008simple,peng-mccallum} .  These models support efficient dynamic-programming inference, but only model {\sl local} dependencies in the output label sequence.  However citations have strong {\sl global} regularities not captured by these models.  For example many book citations contain both an {\sf author} section and an {\sf editor} section, but none have two disjoint {\sf author} sections.  Since linear-chain models are unable to capture more than Markov dependencies, the models sometimes mislabel the {\sl editor} as a second author.  If we could enforce the global constraint that there should be only one {\sf author} section, accuracy could be improved.

One framework for adding such global constraints into tractable models is {\it constrained inference}, in which at inference time the original model is augmented with restrictions on the outputs such that they obey certain global regularities.  When hard constraints can be encoded as linear equations on the output variables, and the underlying model's inference task can be posed as linear optimization, one can formulate this constrained inference problem as an  {\it integer linear program} (ILP) \cite{roth2004linear}.  Alternatively, one can employ {\it dual decomposition} \cite{Rush2010dual}.  Dual decompositions's advantage over ILP is is that it can leverage existing inference algorithms for the original model as a black box.  Such a modular algorithm is easy to implement, and works quite well in practice, providing certificates of optimality for most examples.

The above two approaches have previously been applied to impose {\sl hard} constraints on a model's output.  On the other hand, recent work has demonstrated improvements in citation field extraction by imposing {\sl soft} constraints \cite{ChangRaRo12}.  Here, the model is not required obey the global constraints, but merely pays a penalty for their violation.

This paper introduces a novel method for  imposing soft constraints via dual decomposition.  We also propose a method for learning the penalties the prediction problem incurs for violating these soft constraints.  Because our learning method drives many penalties to zero, it allows practitioners to perform `constraint selection,' in which a large number of automatically-generated candidate global constraints can be considered and automatically culled to a smaller set of useful constraints, which can be run quickly at test time.

Using our new method, we are able to incorporate not only all the soft global constraints of Chang et al. \shortcite{ChangRaRo12}, but also far more complex data-driven constraints, while also providing stronger  optimality certificates than their beam search technique. On a new, more broadly representative, and challenging citation field extraction data set, we show that our methods achieve a 17.9\% reduction in error versus a linear-chain conditional random field. Furthermore, we demonstrate that our inference technique can use and benefit from the constraints of Chang et al. \shortcite{ChangRaRo12}, but that including our data-driven constraints on top of these is beneficial.  While this paper focusses on an application to citation field extraction, the novel methods introduced here would easily generalize to many problems with global output regularities.

      \section{Background}

\subsection{Structured Linear Models}    

The overall modeling technique we employ is to add soft constraints to a simple model for which we have an existing efficient prediction algorithm. For this underlying model, we employ a chain-structured conditional random field  (CRF), since CRFs have been shown to perform better than other simple unconstrained models like hidden markov models for citation extraction \cite{peng-mccallum}. We produce a prediction by performing \textit{MAP inference}~\cite{koller2009probabilistic}. 

The MAP inference task in a CRF be can expressed as an optimization problem with a linear objective~\cite{Sontag_thesis10,SonGloJaa_optbook}. Here, we define a binary indicator variable for each candidate setting
    of each factor in the graphical model. Each of these indicator
    variables is associated with the score that the factor takes on
    when it has the indictor variable's corresponding value. Since the
    log probability of some $y$ in the CRF is proportional to sum of
    the scores of all the factors, we can concatenate the indicator
    variables as a vector $y$ and the scores as a vector $w$ and write the MAP problem as
  \begin{LP}
    \label{eq:general_LP}
      \textbf{max.} & \dt{w,y} \\
      \textbf{s.t.} & y \in \mathcal{U},
    \end{LP}
where the set $\mathcal{U}$ represents the set of valid configurations
of the indicator variables. Here, the constraints are that all neighboring factors agree on the components of $y$ in their overlap. 

\textit{Structured Linear Models} are the general family of models
where prediction requires solving a problem of the
form~\eqref{eq:general_LP}, and they do not always correspond to a
probabilistic model. The algorithms we present in later sections
for handling soft global constraints and for learning the penalties of
these constraints can be applied to general structured linear models,
not just CRFs, provided we have an available algorithm for
performing MAP inference.

    \subsection{Dual Decomposition for Global Constraints}
    \label{sec:dual-decomp-glob}
In order to perform prediction subject to various global constraints, we may need to augment the problem~\eqref{eq:general_LP} with additional constraints. Dual Decomposition is a popular method for performing MAP inference in this scenario, since it leverages  known algorithms for MAP in the base problem where these extra constraints have not been added~\cite{komodakis2007mrf,SonGloJaa_optbook,Rush_JAIR}. In this case, the MAP problem can be formulated as a structured linear model
similar to equation \eqref{eq:general_LP}, for which we have a MAP algorithm,  but
where we have imposed some additional constraints $ Ay
\le b$ that no longer allow us to use the algorithm. In other words,
we consider the problem
    \begin{LP}
      \textbf{max.} & \dt{w,y} \\
      \textbf{s.t.} & y \in \mathcal{U} \\
      & Ay \le b,
    \end{LP}
    for an arbitrary matrix $A$ and vector $b$. We can write the
    Lagrangian of this problem as
    \begin{equation}
      \label{eq:1}
      L(y,\blambda) = \dt{w,y} + \blambda^T(Ay - b). 
    \end{equation}
    Regrouping terms and maximizing over the primal variables, we have
    the dual problem
    \begin{equation}
      \label{eq:2}
      \mathbf{min.}_{\lambda } D(\blambda) = \max_{y \in \mathcal{U}} \dt{w + A^T \blambda, y} -
      \blambda^T b. 
    \end{equation}
For any $\blambda$, we can evaluate the dual objective $D(\lambda)$, since the
maximization in~\eqref{eq:2} is of the same form as the original
problem~\eqref{eq:general_LP}, and we assumed
we had a method for performing MAP in this. Furthermore, a subgradient of $D(\blambda)$ is $A y^* - b$, for an $y^*$ which maximizes this inner optimization problem. Therefore, we
    can minimize $D(\blambda)$ with the projected subgradient method
    \cite{boyd2004convex}, and the optimal $y$ can be obtained when
    evaluating $D(\blambda^*)$. Note that the subgradient of $D(\blambda)$  is the amount by which each constraint is
    violated by $\blambda$ when maximizing over $y$. 

\begin{algorithm}
  \caption{DD: projected subgradient for dual decomposition with hard constraints}
  \label{a:dd-for-joint}
  \begin{algorithmic}[1]
    \begin{small}
      \While{has not converged}
      \State $y^{(t)} = \argmax_{y \in \mathcal{U}} \dt{w + A^T \blambda, y}$ 
      \State $\blambda^{(t)} = \Pi_{0 \le \cdot}\left[\blambda^{(t-1)} - \eta^{(t)} (Ay - b)\right]$
      \EndWhile
    \end{small}
  \end{algorithmic}
\end{algorithm}

    Algorithm~\ref{a:dd-for-joint} depicts the basic projected subgradient
    descent algorithm for dual decomposition. The projection operator
    $\Pi$ consists of truncating all negative coordinates of
    $\blambda$ to 0. This is necessary because $\blambda$ is a vector of dual variables for
    inequality constraints. The algorithm has converged when each constraint is
    either satisfied by $y^{(t)}$ with equality or its corresponding component of $\blambda$ is
    0, due to complimentary slackness~\cite{boyd2004convex}.

\section{Soft Constraints in Dual Decomposition}
    \label{sec:soft-constr-dual}
    We now introduce an extension of Algorithm~\ref{a:dd-for-joint} to
    handle soft constraints. In our formulation, a soft-constrained model imposes a
    penalty for each unsatisfied constraint, proportional to the
    amount by which it is violated. Therefore, our derivation parallels how
    soft-margin SVMs are derived from hard-margin SVMs by introducing
    auxiliary slack variables \cite{cortes1995support}. Note that when performing MAP subject to soft constraints, optimal solutions might not satisfy some constraints, since doing so would reduce
    the model's score by too much. 

    Consider the optimization problems of the form:
    \begin{LP}
      \label{eq:5}
      \textbf{max.} & \dt{w,y} - \dt{c,z} \\
      \textbf{s.t.} & y \in \mathcal{U} \\
      & Ay - b \le z \\
      & -z \le 0,
    \end{LP}
    For positive $c_i$, it is clear that an optimal $z_i$ will be
    equal to the degree to which $a_i^T y \le b_i$ is
    violated. Therefore, we pay a cost $c_i$ times the degree to which
    the $i$th constraint is violated, which mirrors how slack
    variables are used to represent the hinge loss for SVMs. Note that
    $c_i$ has to be positive, otherwise this linear program is
    unbounded and an optimal value can be obtained by setting $z_i$ to
    infinity.

    Using a similar construction as in section \ref{sec:dual-decomp-glob}
    we write the Lagrangian as:
    \begin{dmath}
      \label{eq:3}
      L(y,z,\blambda,\mu) = \dt{w,y} - \dt{c,z} + \blambda^T(Ay - b - z) + \mu^T (-z).
    \end{dmath}
    The optimality constraints with respect to $z$ tell us that $-c -
    \blambda - \mu = 0$, hence $\mu = -c - \blambda$. Substituting, we have
    \begin{equation}
      \label{eq:4}
      L(y,\blambda) = \dt{w,y} + \blambda^T(Ay - b),
    \end{equation}
    except the constraint that $\mu = -c - \blambda$ implies that for
    $\mu$ to be positive $\blambda \le c$. 
    
    Since this Lagrangian has the same form as equation \eqref{eq:1}, we can also derive a dual problem, which is the same as in
    equation \eqref{eq:2}, with the additional constraint that each
    $\blambda_i$ can not be bigger than its cost $c_i$. In other
    words, the dual problem can not penalize the violation of a
    constraint more than the soft constraint model in the primal would penalize you if you violated
    it. 

    This optimization problem
    can still be solved with projected subgradient
    descent and is depicted in Algorithm~\ref{a:soft-dd}. The only modifications to Algorithm~\ref{a:dd-for-joint} are replacing the
    coordinate-wise projection $\Pi_{0 \le \cdot}$ with $\Pi_{0 \le
      \cdot \le c}$ and how we check for convergence. Now, we check for the KKT
    conditions of \eqref{eq:5}, where  for every constraint $i$,
    either the constraint is satisfied with equality, $\blambda_i =
    0$, or $\blambda_i = c_i$.

Therefore, implementing soft-constrained
    dual decomposition is as easy as implementing hard-constrained
    dual decomposition, and the per-iteration complexity is the
    same. We encourage further applications of soft-constraint dual
    decomposition to existing and new NLP problems. 

\begin{algorithm}
  \caption{Soft-DD: projected subgradient for dual decomposition with soft constraints}
  \label{a:soft-dd}
  \begin{algorithmic}[1]
    \begin{small}
      \While{has not converged}
      \State $y^{(t)} = \argmax_{y \in \mathcal{U}} \dt{w + A^T \blambda, y}$ 
      \State $\blambda^{(t)} = \Pi_{0 \le \cdot \le c}\left[\blambda^{(t-1)} - \eta^{(t)} (Ay - b)\right]$
      \EndWhile
    \end{small}
  \end{algorithmic}
\end{algorithm}

    \subsection{Learning Penalties}
    \label{sec:learning-penalties}

    One consideration when using soft v.s. hard constraints is that
soft constraints present a new  training problem, since
    we need to choose the vector $c$, the penalties for violating the
    constraints. 
   An important property of problem~\eqref{eq:5} in the previous
    section is that it corresponds to a
    structured linear model over $y$ and $z$. Therefore, we can apply known training algorithms for
    estimating the parameters of structured linear models to choose
    $c$. 

    All we need to employ the structured perceptron algorithm \cite{collins2002discriminative} or the
    structured SVM algorithm \cite{tsochantaridis2004support} is a
    black-box procedure for performing MAP inference in the structured
    linear model given an arbitrary cost vector. Fortunately, the MAP problem for ~\eqref{eq:5} can be solved using Soft-DD, in Algorithm~\ref{a:soft-dd}.
   
    Each penalty $c_i$  has to be non-negative; otherwise, the
    optimization problem in equation \eqref{eq:5} is ill-defined. This can be ensured by simple modifications of the perceptron and subgradient descent optimization of the structured SVM objective simply by truncating $c$ coordinate-wise to be non-negative at every learning iteration. 

Intuitively, the perceptron update increases the penalty for a constraint if it is satisfied in the ground truth and not in an inferred prediction, and decreases the penalty if the constraint is satisfied in the prediction and not the ground truth. Since we truncate penalties at 0, this suggests that we will learn a penalty of 0 for constraints in three categories: constraints that do not hold in the ground truth, constraints that hold in the ground truth but are satisfied in practice by performing inference in the base CRF model, and constraints that are satisfied in practice as a side-effect of imposing non-zero penalties on some other constraints . A similar analysis holds for the structured SVM approach. 

Therefore, we can view learning the values of the penalties not just as parameter tuning, but as a means to perform `constraint selection,' since constraints that have a penalty of 0 can be ignored. This property allows us to consider large families of constraints, from which the useful ones are automatically identified.

    We found it beneficial, though it is not theoretically necessary,
    to learn the constraints on a held-out development set, separately
    from the other model parameters, as during training most
    constraints are satisfied due to overfitting, which leads to an
    underestimation of the relevant penalties.

   \section{Citation Extraction Data}
     We consider the UMass citation dataset, first introduced in Anzaroot and McCallum~\shortcite{anzaroot2013new}. It has over 1800 citation from many academic fields,
     extracted from the arXiv. This dataset contains both coarse-grained
     and fine-grained labels; for example it contains labels for the
     segment of all authors, segments for each individual author, and
     for the first and last name of each author.  There are 660
     citations in the development set and 367 citation in the test set. 
	 
	 The labels in the UMass dataset are a concatenation of labels from a hierarchically-defined schema. For example, a first name of an author is tagged as: \textit{authors/person/first}. In addition, individual tokens are labeled using a BIO label schema for each level in the hierarchy. 
	 BIO is a commonly used labeling schema for information extraction tasks. BIO labeling allows individual labels on tokens to label segmentation information as well as labels for the segments. In this schema, labels that begin segments are prepended with a \textit{B}, labels that continue a segment are prepended with an \textit{I}, and tokens that don't have a labeling in this schema are given an \textit{O} label. For example, in a hierarchical BIO label schema the first token in the first name for the second author may be labeled as: \textit{I-authors/B-person/B-first}. 
	 
	 An example labeled citation in this dataset can be viewed in figure \ref{example}.
	    
   \section{Global Constraints for Citation Extraction}

     \subsection{Constraint Templates}
	 \label{sec:templates}
         We now describe the families of global constraints we consider for citation extraction. 
     Note these constraints are all
     linear, since they depend only on the counts of each possible
     conditional random field label. Moreover, since our labels are
     BIO-encoded, it is possible, by counting B tags, to count how
     often each citation tag itself appears in a sentence. The first two 
	 families of constraints that we describe are general to any sequence 
	 labeling task while the last is specific to hierarchical labeling 
	 such as available in the UMass dataset.  
     
     Our sequence output is denoted as $y$ and an element of this sequence is $y_k$. 
     
	We denote $\left[ \left[ y_k = i \right]\right]$ as the function
     that outputs $1$ if $y_k$ has a $1$ at index $i$ and $0$ otherwise. 
	 Here, $y_k$ represents an output tag of the CRF, so if $\left[ \left[ y_k = i \right]\right]$ = 1, then we have that $y_k$ was given a label with index $i$.

	 \subsection{Singleton Constraints}
	 Singleton constraints ensure that each label can appear at most once in a citation. These are same global constraints that were used for citation field extraction in Chang et al.~\shortcite{ChangRaRo12}. We define $s(i)$ to be the number of times the label with index $i$ is predicted in a citation, formally: 
	 
	 \[ s(i) = \sum_{y_k \in y} \left[\left[ y_k = i \right]  \right] \]
	 
	 The constraint that each label can appear at most once takes the form:
	 
	 \[ s(i) <= 1 \]
	 	 
	 \subsection{Pairwise Constraints}
	 
	 Pairwise constraints are constraints on the counts of two labels in a citation. We define $z_1(i,j)$ to be  
	 
	 \[ z_1(i,j) = \sum_{y_k \in y} \left[\left[ y_k = i \right] \right] + \sum_{y_k \in y} \left[\left[ y_k = j \right] \right] \]
	 
	 and $z_2(i,j)$ to be 
	 
	 \[ z_2(i,j) = \sum_{y_k \in y} \left[\left[ y_k = i \right] \right] - \sum_{y_k \in y} \left[\left[ y_k = j \right] \right] \]
	 
	We consider all constraints of the forms: $ z(i,j) \le
        {0,1,2,3}$ and $ z(i,j) \ge {0,1,2,3}$. 
         
Note that some pairs of these constraints are redundant or logically incompatible. However, we are using them as soft constraints, so these constraints will not necessarily be satisfied by the output of the model, which eliminates concern over enforcing logically impossible outputs. Furthermore, in section~\ref{sec:learning-penalties} we described how our procedure for learning penalties will drive some penalties to 0, which effectively removes them from our set of constraints we consider. It can be shown, for example, that we will never learn non-zero penalties for certain pairs of logically incompatible constraints using the perceptron-style algorithm described in section~\ref{sec:learning-penalties}  . 
	 
 	 \begin{figure*}
 		 \small
 		 \begin{tabular}{l|lp{10cm}}
 		 \hline
			 
 		 Unconstrained & \parbox{13.2cm}{\vspace{.5\baselineskip} \segments{[17]}{ref-marker} \segment{ \segmentm{ \segments{D.}{first} \segments{Sivia ,}{last} }{person} \segmentm{ \segments{J.}{first} \segments{Skilling ,}{last} }{person} }{authors} \segment{ \segmentm{ Data Analysis : A Bayesian Tutorial ,}{\color{red}{booktitle}} \segmentm{ Oxford University Press , }{publisher} \segmentm{ \segments{ 2006 }{year} }{date} }{venue} \vspace{.5\baselineskip} }\\
 		 \hline
 		 Constrained  & \parbox{13.2cm}{\vspace{.5\baselineskip} \segments{[17]}{ref-marker} \segment{ \segmentm{ \segments{D.}{first} \segments{Sivia ,}{last} }{person} \segmentm{ \segments{J.}{first} \segments{Skilling ,}{last} }{person} }{authors} \segmentm{ Data Analysis : A Bayesian Tutorial ,}{\color{red}{title}} \segment{ \segmentm{ Oxford University Press , }{publisher} \segmentm{ \segments{ 2006 }{year} }{date} }{venue} \vspace{.5\baselineskip} }\\
 		 \hline
 		 \end{tabular}
		 \begin{tabular}{l|lp{10cm}}
			 \hline
			 Unconstrained & \parbox{13.2cm}{\vspace{.5\baselineskip} \segment{ \segmentm{ \segments{Sobol' ,}{last} \segments{I.}{first} \segments{M.}{middle} }{person} }{authors} \segment{ \segmentm{(1990) .}{year} }{date} \segment{On sensitivity estimation for nonlinear mathematical models .}{title} \segment{ \segmentm{Matematicheskoe Modelirovanie ,}{journal} \segmentm{2}{volume} \segmentm{(1) :}{number} \segmentm{112--118 .}{pages} \segmentm{ ( In Russian ) . }{{\color{red} status}} \vspace{.5\baselineskip}}{venue}}\\
			 \hline
			 Constrained & \parbox{13.2cm}{\vspace{.5\baselineskip} \segment{ \segmentm{ \segments{Sobol' ,}{last} \segments{I.}{first} \segments{M.}{middle} }{person} }{authors} \segment{ \segmentm{(1990) .}{year} }{date} \segment{On sensitivity estimation for nonlinear mathematical models .}{title} \segment{ \segmentm{Matematicheskoe Modelirovanie ,}{journal} \segmentm{2}{volume} \segmentm{(1) :}{number} \segmentm{112--118 .}{pages} \segmentm{ ( In Russian ) . }{{\color{red} language}} \vspace{.5\baselineskip} }{venue}}\\
		 \hline
		 \end{tabular}
 		\caption{ Two examples where imposing soft global constraints improves field extraction errors. Soft-DD converged in 1 iteration on the first example, and 7 iterations on the second. 
		When a reference is citing a book and not a section of the book, the correct labeling of the name of the book is {\sf title}. In the first example, the baseline CRF incorrectly outputs {\sf booktitle}, but this is fixed by Soft-DD, which penalizes outputs based on the constraint that {\sf booktitle} should co-occur with an {\sf address} label. 
		In the second example, the unconstrained CRF output violates the constraint that {\sf title} and {\sf status} labels should not co-occur. The ground truth labeling also violates a constraint that {\sf title} and {\sf language} labels should not co-occur. At convergence of the Soft-DD algorithm, the correct labeling of {\sf language} is predicted, which is possible because of the use of soft constraints. }
 		 
 	    \label{sub-example}
	 \end{figure*}
	 
	 \subsection{Hierarchical Equality Constraints}
	 
	 The labels in the citation dataset are hierarchical labels. This means that the labels are the concatenation of all the levels in the hierarchy. We can create constraints that are dependent on only one or couple of elements in the hierarchy. 
	 
	 We define $C(x,i)$ as the function that returns 1 if the output $x$ contains the label $i$ in the hierarchy and 0 otherwise. We define $e(i,j)$ to be
	 
	 \[ e(i,j) = \sum_{y_k \in y} \left[\left[ C(y_k,i) \right]  \right] - \sum_{y_k \in y} \left[\left[ C(y_k,j) \right]  \right] \]
	 
	 Hierarchical equality constraints take the forms:
         \begin{eqnarray}
           \label{eq:7}
           e(i,j) &\ge& 0\\
           e(i,j) &\le& 0 
         \end{eqnarray}

	 \subsection{Local constraints}
	 
	 We constrain the output labeling of the chain-structured CRF to be a valid BIO encoding. This both improves performance of the underlying model when used without global constraints, as well as ensures the validity of the global constraints we impose, since they operate only on  \textit{B} labels. The constraint that the labeling is valid BIO can be expressed as a collection of pairwise constraints on adjacent labels in the sequence. Rather than enforcing these constraints using dual decomposition, they can be enforced directly when performing MAP inference in the CRF by modifying the dynamic program of the Viterbi algorithm to only allow valid pairs of adjacent labels. 
	 
	 \begin{table}
	 \centering
	 \small
	 \begin{tabular}{|l|l|l|l|}
	 \hline
	 Constraints & F1 score & Sparsity & \# of cons\\
	 \hline
	 \hline
	 Baseline & 94.44 & & \\
	 \hline
	 Only-one & 94.62 & 0\% & 3 \\
	 \hline
	 Hierarchical & 94.55 & 56.25\% & 16\\
	 \hline
	 Pairwise & 95.23 & 43.19\% & 609\\
	 \hline
	 All & \textbf{95.39} & 32.96\% & 628\\
	 \hline
	 \hline
	 All DD & 94.60 & 0\% & 628\\
	 \hline
	 \end{tabular}
	 \caption{Set of constraints learned and F1 scores. The last row depicts the result of inference using all constraints as hard constraints.}
	 \label{constraints}
	 \end{table}

     \subsection{Constraint Pruning}
	\label{sec:pruning} 
     While the techniques from section \ref{sec:learning-penalties}
     can easily cope with a large numbers of constraints at training
     time, this can be computationally costly, specially if one is
     considering very large constraint families. This is problematic because the size of some constraint families we consider grows quadratically with the number of candidate labels, and there are about 100 in the UMass dataset. Such a family consists of constraints that the sum of the
     counts of two different label types has to be bounded (a useful
     example is that there can't be more than one out of ``phd
     thesis'' and ``journal''). Therefore, quickly pruning bad constraints can save a  substantial amount of training time, and can lead to better generalization. 
	 
     To do so, we calculate a score that estimates how useful each
     constraint is expected to be. Our score compares how often the
     constraint is violated in the ground truth examples versus our predictions. Here, prediction is done with respect to the base chain-structured CRF tagger and does not include global constraints. 
     Note that it may make sense to consider a
     constraint that is sometimes violated in the ground truth, as the penalty learning
     algorithm can learn a small penalty for it, which will allow it to be
     violated some of the time. Our importance score is defined as, for
     each constraint $c$ on labeled set $D$,
     \begin{align}
       imp(c) = \frac{\sum_{d \in D} [[max_{y} w_d^T y]]_c }{\sum_{d \in D} [[y_d]]_c},
     \end{align}
     where $[[y]]_c$ is 1 if the constraint is violated on output $y$ and 0 otherwise. Here, $y_d$ denotes the ground truth labeling and $w_d$ is the vector of scores for the CRF tagger. 
     
     We prune constraints by picking a cutoff value for $imp(c)$. A
     value of $imp(c)$ above 1 implies that the constraint is more
     violated on the predicted examples than on the ground truth, and
     hence that we might want to keep it.
     
     We also find that the constraints that have the largest $imp$
     values are semantically interesting.

 	\section{Related Work}

There are multiple previous examples of augmenting chain-structured sequence models with terms capturing global relationships by expanding the chain to a more complex graphical model with non-local dependencies between the outputs. Inference in these models can be performed, for example,  with loopy belief propagation \cite{bunescu2004collective,sutton2004collective} or Gibbs sampling~\cite{finkel2005incorporating}. Belief propagation is prohibitively expensive in our model due to the high cardinalities of the output variables and of the global factors, which involve all output variables simultaneously. There are various methods for exploiting the combinatorial structure of these factors, but performance would still have higher complexity than our method. While Gibbs sampling has been shown to work well tasks such as named entity recognition~\cite{finkel2005incorporating}, our previous experiments show that it does not work well for citation extraction, where it found only  low-quality solutions in practice because the sampling did not mix well, even on a simple chain-structured CRF. 

Recently, dual decomposition has become a popular method for solving complex structured prediction problems in NLP \cite{koo2010dual,Rush2010dual,Rush_JAIR,paul2012implicitly,chieu2012combining}. Soft constraints can be implemented inefficiently using hard constraints and dual decomposition--- by introducing copies of output variables and an auxiliary graphical model, as in Rush et al.~\shortcite{rush2012improved}. However, at every iteration of dual decomposition, MAP must be run in this auxiliary model. Furthermore the copying of variables doubles the number of iterations needed for information to flow between output variables, and thus slows convergence. On the other hand, our approach to soft constraints has identical per-iteration complexity as for hard constraints, and is a very easy modification to existing hard constraint code. 

	 Initial work in machine learning for citation extraction used Markov models with no global constraints. Hidden Markov models (HMMs), were originally employed for automatically extracting information from research papers on the CORA dataset \cite{seymore1999learning,hetzner2008simple}. Later, CRFs were shown to perform better on CORA, improving the results from the Hmm's token-level F1 of 86.6 to 91.5 with a CRF\cite{peng-mccallum}.

 	Recent work on globally-constrained inference in citation extraction used an HMM$^{CCM}$, which is an HMM with the addition of global features that are restricted to have positive weights \cite{ChangRaRo12}. Approximate inference is performed using beam search. This method increased the HMM token-level accuracy from 86.69 to 93.92 on a test set of 100 citations from the CORA dataset. The global constraints added into the model are simply that each label only occurs once per citation. This approach is limited in its use of an HMM as an underlying model, as it has been shown that CRFs perform significantly better, achieving 95.37 token-level accuracy on CORA \cite{peng-mccallum}. In our experiments, we demonstrate that the specific global constraints used by Chang et al.~\shortcite{ChangRaRo12} help on the UMass dataset as well. 

	 \section{Experimental Results}	 
	 Our baseline is the one used in Anzaroot and McCallum~\shortcite{anzaroot2013new}, with some labeling errors removed. This is a chain-structured CRF trained to maximize the conditional likelihood using L-BFGS with L2 regularization. 
	 
	 We use the same features as Anzaroot and McCallum~\shortcite{anzaroot2013new}, which include word type, capitalization, binned location in citation, regular expression matches, and matches into lexicons. In addition, we use a rule-based segmenter that segments the citation string based on punctuation as well as probable start or end segment words (e.g. `in' and `volume'). We add a binary feature to tokens that correspond to the start of a segment in the output of this simple segmenter. This final feature improves the F1 score on the cleaned test set from 94.0 F1 to 94.44 F1, which we use as a baseline score. 
	 
We then use the development set to learn the penalties for the soft constraints, using the perceptron algorithm described in section \ref{sec:learning-penalties}. MAP inference in the model with soft constraints is performed using Soft-DD, shown in Algorithm~\ref{a:soft-dd}.
	 
	 We instantiate constraints from each template in section~\ref{sec:templates},  iterating over all possible labels that contain a \textit{B} prefix at any level in the hierarchy and pruning all constraints with $imp(c) < 2.75$ calculated on the development set. We asses performance in terms of field-level F1 score, which is the harmonic mean of precision and recall for predicted segments. 

Table \ref{constraints} shows how each type of constraint family improved the F1 score on the dataset. Learning all the constraints jointly provides the largest improvement in F1 at 95.39. This improvement in F1 over the baseline CRF as well as the improvement in F1 over using only-one constraints was shown to be statistically significant using the Wilcoxon signed rank test with p-values $< 0.05$. In the all-constraints settings, 32.96\% of the constraints have a learned parameter of $0$, and therefore only 421 constraints are active. 
	 Soft-DD converges, and thus solves the constrained inference problem exactly, for all test set examples after at most 41 iterations. Running Soft-DD to convergence requires 1.83 iterations on average per example. Since performing inference in the CRF is by far the most computationally intensive step in the iterative algorithm, this means our procedure requires approximately twice as much work as running the baseline CRF on the dataset. On examples where unconstrained inference does not satisfy the constraints, Soft-DD converges after 4.52 iterations on average. For 11.99\% of the examples, the Soft-DD algorithm satisfies constraints that were not satisfied during unconstrained inference, while in the remaining 11.72\% Soft-DD converges with some constraints left unsatisfied, which is possible since we are imposing them as soft constraints. 
	 
	 We could have enforced these constraints as hard constraints rather than soft ones. This experiment is shown in the last row of Table \ref{constraints}, where F1 only improves to 94.6. In addition, running the DD algorithm with these constraints takes 5.21 iterations on average per example, which is 2.8 times slower than Soft-DD with learned penalties.

In Figure \ref{earlystopping}, we analyze the performance of Soft-DD when we don't necessarily run it to convergence, but stop after a fixed number of iterations on each test set example. We find that a large portion of our gain in accuracy can be obtained when we allow ourselves as few as 2 dual decomposition iterations. However, this only amounts to 1.24 times as much work as running the baseline CRF on the dataset, since the constraints are satisfied immediately for many examples. 

	 In Figure \ref{sub-example} we consider two applications of our Soft-DD algorithm, and provide analysis in the caption. 
	 	 	 	 
	 \begin{table}
	 \centering	 
	 \small
	 \begin{tabular}{|l|lll|}
	 \hline
	 Stop & F1 score & Convergence & Avg Iterations  \\
	 \hline
	 1 & 94.44 & 76.29\% & 1.0\\
	 \hline
	 2 & 95.07 & 83.38\% & 1.24\\
	 \hline
	 5 & 95.12 & 95.91\% & 1.61\\
	 \hline
	 10 & 95.39 & 99.18\% & 1.73\\
	 \hline
	 \end{tabular}
         \caption{Performance from terminating Soft-DD early. Column 1 is the number of iterations we allow each example. Column 3 is the \% of test examples that converged. Column 4 is the average number of necessary iterations, a surrogate for the slowdown over performing unconstrained inference.  }
	 \label{earlystopping}
	 \end{table}

	 We train and evaluate on the UMass dataset instead of CORA, because it is significantly larger, has a useful finer-grained labeling schema, and its annotation is more consistent. We were able to obtain better performance on CORA using our baseline CRF than the $HMM^{CCM}$ results presented in Chang et al.~\shortcite{ChangRaRo12}, which include soft constraints. Given this high performance of our base model on CORA, we did not apply our Soft-DD algorithm to the dataset. Furthermore, since the dataset is so small, learning the penalties for our large collection of constraints is difficult, and test set results are unreliable. Rather than compare our work to Chang et al.~\shortcite{ChangRaRo12} via results on CORA, we apply their constraints on the UMass data using Soft-DD and demonstrate accuracy gains, as discussed above. 
	 
	 \subsection{Examples of learned constraints}
	 We now describe a number of the useful constraints that receive non-zero learned penalties and have high importance scores, defined in Section~\ref{sec:pruning}. The importance score of a constraint provides information about how often it is violated by the CRF, but holds in the ground truth, and a non-zero penalty implies we enforce it as a soft constraint at test time.
	 
	 \label{sec:example-of-constraints}
	 
	 The two singleton constraints with highest importance score are that there should only be at most one {\sf title} segment in a citation and that there should be at most one {\sf author} segment in a citation. The only one {\sf author} constraint is particularly useful for correctly labeling {\sf editor} segments in cases where  unconstrained inference mislabels them as {\sf author} segments. As can be seen in Table ~\ref{highest}, {\sf editor} fields are among the most improved with our new method, largely due to this constraint.
	 
	 The two hierarchical constraints with the highest importance scores with non-zero learned penalties constrain the output such that number of {\sf person} segments does not exceed the number of {\sf first} segments and vice-versa. Together, these constraints penalize outputs in which the number of {\sf person} segments do not equal the number of {\sf first} segments, i.e., every author should have a first name.
	 
	 One important pairwise constraint penalizes outputs in which {\sf thesis} segments don't co-occur with {\sf school} segments. {\sf School} segments label the name of the university that the thesis was submitted to. The application of this constraint increases the performance of the model on {\sf school} segments dramatically, as can be seen in table \ref{highest}.
	 
	 An interesting form of pairwise constraints penalize outputs in which some labels do not co-occur with other labels. Some examples of constraints in this form enforce that {\sf journal} segments should co-occur with {\sf pages} segments and that {\sf booktitle} segments should co-occur with {\sf address} segments. An example of the latter constraint being employed during inference is the first example in Figure \ref{sub-example}. Here, the constrained inference penalizes output which contains a {\sf booktitle} segment but no {\sf address} segment. This penalization leads allows the constrained inference to correctly label the {\sf booktitle} segment as a {\sf title} segment.
	 
	 The above example constraints are almost always satisfied on the ground truth, and would be useful to enforce as hard constraints. However, there are a number of learned constraints that are often violated on the ground truth but are still useful as soft constraints. 
Take, for example, the constraint that the number of {\sf number} segments does not exceed the number of {\sf booktitle} segments, as well as the constraint that it does not exceed the number of {\sf journal} segments. These constraints are moderately violated on ground truth examples, however. For example, when {\sf booktitle} segments co-occur with {\sf number} segments but not with {\sf journal} segments, the second constraint is violated. It is still useful to impose these soft constraints, as strong evidence from the CRF allows us to violate them, and they can guide the model to good predictions when the CRF is unconfident. 
		 	 \begin{table}
		 		 \small
		  	\begin{tabular}{llll}
		 	  Label & U & C & + \\
		 	  \hline
		 	  \hline
		 		venue/series & 35.29 & 66.67 & 31.37\\
		 		venue/editor/person/first & 66.67 & 94.74 & 28.07\\
		 		venue/school & 40.00 & 66.67 & 26.67\\
		 		venue/editor/person/last & 75.00 & 94.74 & 19.74\\
		 		venue/editor & 77.78 & 90.00 & 12.22\\
		 		venue/editor/person/middle & 81.82 & 91.67 & 9.85\\
		 		\hline
		 	\end{tabular}
		 	\caption{Labels with highest improvement in F1. U is in unconstrained inference. C is the results of constrained inference. + is the improvement in F1.}
		 	\label{highest}
		 	\end{table}
	 	 	 
	  \section{Conclusion}
	  We introduce a novel modification to the standard projected subgradient dual decomposition algorithm for performing MAP inference subject to hard constraints to one for performing MAP in the presence of soft constraints.  In addition, we offer an easy-to-implement procedure for learning the penalties on soft constraints. This method drives many penalties to zero, which allows users to automatically discover discriminative constraints from large families of candidates. 
	  
         We show via experiments on a recent substantial dataset that using soft constraints, and selecting which constraints to use with our penalty-learning procedure, can lead to significant gains in accuracy. We achieve a 17\% gain in accuracy over a chain-structured CRF model, while only needing to run MAP  in the CRF an average of less than 2 times per example. This minor incremental cost over Viterbi, plus the fact that we obtain certificates of optimality on 100\% of our test examples in practice, suggests the usefulness of our algorithm for large-scale applications. We encourage further use of our Soft-DD procedure for other structured prediction problems.  
	 	 
	 \section*{Acknowledgments}
	 This work was supported in part by the Center for Intelligent Information Retrieval, in part by DARPA under agreement number FA8750-13-2-0020, in part by NSF grant \#CNS-0958392, and in part by IARPA via DoI/NBC contract \#D11PC20152. The U.S. Government is authorized to reproduce and distribute reprint for Governmental purposes notwithstanding any copyright annotation thereon. Any opinions, findings and conclusions or recommendations expressed in this material are those of the authors and do not necessarily reflect those of the sponsor.

\bibliographystyle{acl}
\bibliography{CitationPaper}

\end{document}